\documentclass[letterpaper, 10 pt, conference]{ieeeconf}  

\IEEEoverridecommandlockouts                              

\usepackage{times}
\usepackage{epsfig}
\usepackage{graphicx}
\usepackage{amsmath}
\usepackage{amssymb}
\usepackage{algorithm}
\usepackage[noend]{algpseudocode}
\usepackage{lipsum}

\usepackage{amsmath}
\usepackage{array, makecell} 
\usepackage{subcaption}
\DeclareCaptionSubType*[Alph]{table}
\DeclareCaptionLabelFormat{mystyle}{Table~\bothIfFirst{#1}{ ̃}#2}
\usepackage{rotating}

\usepackage{color}
\usepackage{xcolor}
\usepackage{comment}
\usepackage{graphicx}
\usepackage{booktabs}
\usepackage[margin=2.5cm]{geometry}
\usepackage{caption}

\title{\LARGE \bf
Pedestrian Collision Avoidance for Autonomous Vehicles at Unsignalized Intersection Using Deep Q-Network}

\author{Kasra Mokhtari$^{1}$ and Alan R. Wagner$^{2}$
\thanks{$^{1}$Department of Mechanical Engineering,
        The Pennsylvania State University, State College, PA 16802, USA
        {\tt\small kbm5402@psu.edu}}%
\thanks{$^{2}$Department of Aerospace Engineering, The Pennsylvania State University,
        State College, PA 16802, USA
        {\tt\small alan.r.wagner@psu.edu}}%
}

\begin{document}

\maketitle
\thispagestyle{empty}
\pagestyle{empty}

\begin{abstract}
\label{Abstract}
Prior research has extensively explored Autonomous Vehicle (AV) navigation in the presence of other vehicles, however, navigation among pedestrians, who are the most vulnerable element in urban environments, has been less examined. This paper explores AV navigation in crowded, unsignalized intersections. We compare the performance of different deep reinforcement learning methods trained on our reward function and state representation. The performance of these methods and a standard rule-based approach were evaluated in two ways, first at the unsignalized intersection on which the methods were trained, and secondly at an unknown unsignalized intersection with a different topology. For both scenarios, the rule-based method achieves less than 40\% collision-free episodes,  whereas our  methods result in a performance of approximately 100\%. Of the three methods used, DDQN/PER  outperforms the other two methods while it also shows the smallest average intersection crossing time, the greatest average speed, and the greatest distance from the closest pedestrian.
\end{abstract}
 
\section{Introduction}
\label{sec1}
According to the U.S. Federal Highway Administration, approximately 40\% of car accidents and almost 70\% of fatalities occur at unsignalized intersections~\cite{coakley2009intersection}. Driving through urban unsignalized intersections without the guidance of centralized traffic lights and traffic signs is a challenging task for autonomous vehicle (AV). In these scenarios, an AV must autonomously decide when and how to navigate an intersection safely and efficiently, accounting for the intention of pedestrians, cyclists, and other cars. 

\begin{figure}[t]
    \centering
    \includegraphics[scale = 0.55]{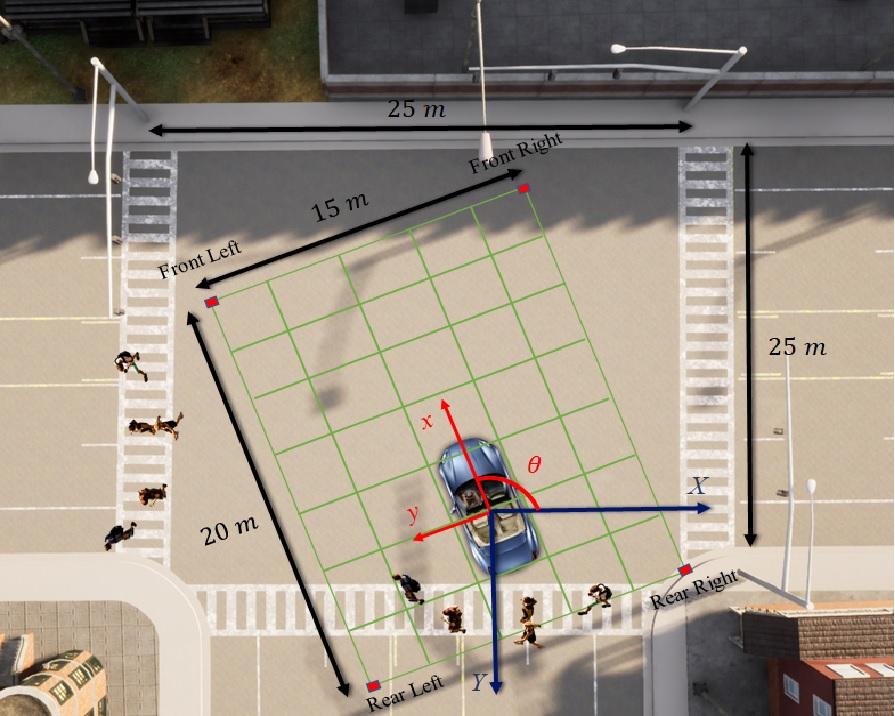}
    \caption{A bird's eye view of unsignalized intersection is depicted. The global axes are presented with the blue arrows. The region of interest (ROI) is shown by the green grid. For better visualization, the grid is displayed as larger than was used in the simulation (best viewed in color).}
    \label{fig1:intersection}
\end{figure}

Reinforcement learning (RL) combined with deep learning has achieved outstanding success in various areas such as video games~\cite{torrado2018deep} and robotics~\cite{kober2013reinforcement}. These advances have inspired the research community to examine the performance of deep reinforcement learning in autonomous driving~\cite{tram2018learning}. Although, prior research has extensively explored AV navigation in the presence of other vehicles at unsignalized intersections~\cite{isele2018navigating, isele2018safe, liu2020decision}, navigating amongst pedestrians, who are the most vulnerable element in the urban environment, is under explored.

This paper explores the potential use of deep reinforcement learning methods as a means to design a real-time controller for an AV navigation through an unsignalized intersection crowded with pedestrians in a high-fidelity simulation such as CARLA. We compare three potential methods for deep reinforcement learning: deep Q Networks (DQN)~\cite{mnih2015human}, Double DQNs (DDQN)~\cite{van2015deep}, and DDQNs integrated with prioritized experience replay (DDQN/PER)~\cite{schaul2015prioritized}. Our environment is depicted in Figure~\ref{fig1:intersection} using CARLA, an open-source simulator. To the best of our knowledge, using the 3D state space representation and our unique conditional reward function, we have developed the first decision-making process for autonomous vehicle navigation among crowds that offers safe behavior ($100\%$ collision-free episodes) for an AV's navigation in a pedestrians-rich urban setup tested in a high-fidelity autonomous driving simulation. Moreover, our approach can handle different numbers of pedestrians without a noticeable increase in the computation time which makes it more suitable for realistic autonomous driving scenarios.

The remainder of this paper is organized as follows: Section~\ref{sec2} presents related work, Section~\ref{sec3} presents the Markov decision process (MDP) framework and deep RL (DRL) algorithms for three different methods. The problem statement and the proposed DRL implementation are discussed in Section~\ref{sec4}. Section~\ref{sec5} introduces the simulation setup and the experiment on which we examine the performance of the RL approaches. Section~\ref{sec6} demonstrates the performance evaluation results, and finally Section~\ref{sec7} offers conclusions and directions for future work. 
 
\section{Related Work}
\label{sec2}

Navigation through crowds using small ground vehicles has been well studied. Museum tour guide robots, for example, detect and navigate around people who block the robot's path~\cite{thrun1999minerva}. The pedestrian dominance model (PDM) proposed in~\cite{randhavane2019pedestrian} is capable of navigating through groups of pedestrians by identifying the dominant characteristics of these groups based on their trajectories. The human-aware motion planner (HAMP) proposed in~\cite{sisbot2007human}, considers the safety of the robot's movement as well as the human's comfort while attempting to keep the robot in front of people and visible at all times. A constraint-optimizing method for person–acceptable navigation (COMPANION)~\cite{kirby2010social} models human social conventions, including avoiding people's personal space, as well as task-based constraints such as minimizing distance.

Although these model-based approaches implemented on ground robots are capable of navigating through groups of pedestrians, they tend to be computationally expensive and hard to adapt to higher dimensional autonomous driving problems where there are multiple road users in the environment. As an alternative, a model for negotiating between an AV and a pedestrian at an unsignalized intersection using discrete sequential game theory was developed in~\cite{fox2018should}. However, game-theory models are overly simplified while attempting to capture complex interactions of AVs and pedestrians. A partially observable Markov decision process (POMDP) has also been used to model the interactions between pedestrians and AVs at intersections~\cite{barbier2018probabilistic}. Yet POMDP algorithms are generally intractable preventing their application to different conditions such as scenarios with additional road users such as other vehicles and pedestrians. One of the most widely used approaches to address the pedestrian collision avoidance problem for an AV is a rule-based method built on time-to-collision (TTC). The rule-based method identifies and utilizes a set of relational rules that collectively represents the contextual knowledge captured by the system. While the rule-based method is relatively reliable and easily interpretable, it has some limitations such as requiring full knowledge of the environment, and being overly cautious by assuming a constant velocity for other cars and pedestrians. 

Reinforcement learning (RL) is a promising approach for AV navigation around pedestrians. Deshpande et al. explore the use of deep reinforcement learning for navigation among
pedestrians by training a deep Q-network (DQN) based on simulation data to cross an intersection with pedestrians present~\cite{deshpande2019deep}. They assume a grid-based state-space representation of the environment and an overly simplified pedestrian model. Unfortunately, the learned policy does not guarantee pedestrian safety. The concept of POMDP planning and reinforcement learning are integrated in~\cite{bouton2019safe} to develop efficient policies with probabilistic safety assurances while also including the AV’s interactions with pedestrians and other vehicles. Uncertainty about the other road users’ (e.g., pedestrians, cyclists, and other vehicles) course of action is captured by the transition model and state uncertainty. Although this RL method guarantees safe navigation of the AV around the pedestrians, the computation time drastically grows with the number of existing vehicles and pedestrians in the environment (due to applying the scene decomposition algorithm on every road user). Therefore, this approach is not suitable for real-time autonomous driving applications. 
\section{Technical Background}
\label{sec3}
\subsection{Reinforcement Learning}
\label{sec3.A}

A reinforcement learning task is typically formulated as a Markov decision process (MDP) defined by the 5-tuple $(S, A, P, R)$, characterized by a set of states $S$, a set of actions $A$, state transition probability $P$, and a scalar reward function $R$~\cite{sutton2018reinforcement}. In a reinforcement learning framework, at each time step $t$, the agent takes an action $a_t$ based on a policy $\pi(a|s)$, it receives a scalar reward $r_t$ and transitions into the next state $s_{t+1}$~\cite{kaelbling1996reinforcement}. This interaction continues until the agent reaches a terminal state, at which point it restarts. The goal of the agent is to maximize the expected discounted reward $R_t=\sum_{t=0}^T \gamma^k r_{t}$, where a discount factor $\gamma \in (0,1]$ emphasizes the importance of the future reward.

To tackle this optimization problem, Q-learning method is employed where actions are selected using state-value function $Q(s,a)$~\cite{watkins1992q}. The optimal action $a^{'}$ for a state $s$ is computed as:
\begin{equation}
\label{eq.y_DQN}
a^{'}=arg\max_{a\in A}Q(s,a)
\end{equation}

\noindent In Deep Q-Network (DQN), the $Q(s,a)$ is approximated using a deep neural network (called online network)~\cite{mnih2015human}. A Double Deep Q-Network is a variant of DQN, where a separate target network for value estimation is used to make the learning process more stable~\cite{van2015deep}. The target network parameters are periodically updated using the latest parameter of the online network. 

To train the DQN and DDQN more effectively, replay memory is employed in which the transition sequences are stored in a replay buffer as the 4-tuple of $(s_t, a_t, r_t, s_{t+1})$ at every time step~\cite{lin1992self}. The transition includes the current state, the selected action, the corresponding received reward, and the subsequent state. Therefore, transitions are uniformly sampled from replay memory, regardless of their significance. However, transitions might vary in their task relevance, and therefore, are given different priorities when training the agent. Prioritized experience replay (PER) assigns different sampling weights to each transition based on the calculated Temporal Difference (TD) error~\cite{schaul2015prioritized}. The probability of sampling transition $i$ is then defined as:
\begin{equation}
P(i) = \frac{p_{i}^{\alpha}}{\sum_{k}{p_{k}^{\alpha}}}
\end{equation}
where $p(i) > 0 $ is the priority of transition $i$, k is the total number of transitions in the replay memory and $\alpha$ determines how much priority is desired (with $\alpha = 0$ corresponding to uniform random sampling). In proportional prioritization, $p(i)=\delta_{i}+\epsilon$, where $\epsilon$ is a small positive constant. Importance sampling is applied to avoid the bias in the updated distribution and the importance-sampling (IS) weights are computed as:
\begin{equation}
w(i) = (\frac{1}{N} \times \frac{1}{P(i)})^\beta
\end{equation}
where $N$ is the replay memory size, and $\beta$ determines the compensation degree that fully compensates for the non-uniform probabilities $P(i)$ if $\beta = 1$. In practice, $\beta$ linearly anneals from its initial value $\beta_{0}$ to $1$. 
\section{Methodology}
\label{sec4}

\begin{figure*}[t]
    \centering
    \includegraphics[scale = 0.45]{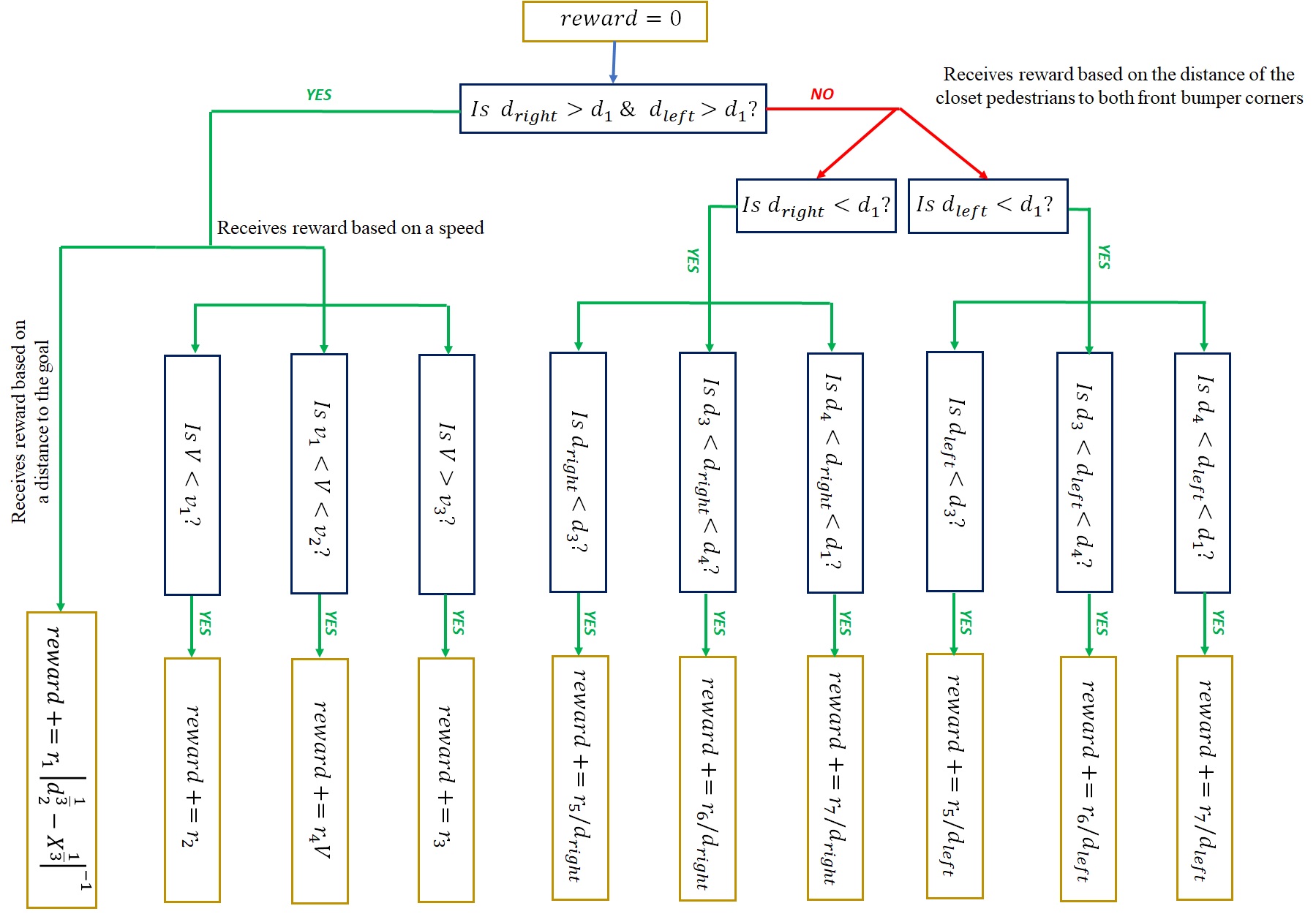}
    \caption{A reward Function flow chart.}
    \label{fig2:rewardfunction}
\end{figure*}

\subsection{Problem Statement}
\label{sec4.A}

We investigate the problem of Autonomous Vehicle navigation at unsignalized intersection in a structured pedestrian-rich urban setup. A three-way intersection scenario is considered where the ego vehicle's goal is to make a left-turn without colliding with pedestrians or violating the speed limit. It is assumed that the environment is fully observable indicating that pedestrian information collected by sensors (such as LIDAR) is accurate and sufficient to make a decision at every time step. The decision-making process is modeled as an MDP.  

\subsection{MDP Formulation}
\label{sec4.B}
\subsubsection{State Space}
\label{sec4.B.1}
We use a similar approach to~\cite{deshpande2019deep}, where the grid-based state space representation is in the form of a 3-D tensor which is used to describe the state of the ego vehicle and the pedestrians. A region of interest (ROI) of the environment around the ego vehicle with the length $L$ and the width $W$ is discretized into multiple grids each with the grid discretization $l \times w$ as shown in Figure~\ref{fig1:intersection}. The ROI is constructed in such a way that the ego vehicle's center of gravity always lies on a cell with an index ($\frac{4L}{5l}$,$\frac{W}{2w}$) and the ROI's orientation remains the same with respect to the ego vehicle direction. Therefore, given the ego vehicle location in the global coordinate system $(EV_x, EV_y)$ and ego vehicle direction $EV_\theta$, the ROI vertices Cartesian coordinates are shown in Table~\ref{tab1:coordinates}. 

\begin{table}
\begin{center}
\scalebox{0.9}{
\begin{tabular}{ |c|c|c| } 
\hline
Vertices & Cartesian Coordinates \\
\hline
Rear Right x  & $EV_x+ \frac{L\times cos(\theta)}{5}+\frac{W \times sin(\theta)}{2}$\\ 
Rear Right y  & $EV_y+ \frac{L\times sin(\theta)}{5}-\frac{W \times cos(\theta)}{2}$\\ 
Rear Left x   & $EV_x+ \frac{L\times cos(\theta)}{5}-\frac{W \times sin(\theta)}{2}$\\ 
Rear Left y   & $EV_y+ \frac{L\times sin(\theta)}{5}+\frac{W \times cos(\theta)}{2}$\\ 
Front Right x & $EV_x- \frac{4L\times cos(\theta)}{5}+\frac{W \times sin(\theta)}{2}$\\ 
Front Right y & $EV_y- \frac{4L\times sin(\theta)}{5}-\frac{W \times cos(\theta)}{2}$\\ 
Front Left x  & $EV_x - \frac{L\times cos(\theta)}{5}-\frac{W \times sin(\theta)}{2}$\\ 
Front Left y  & $EV_y - \frac{L\times sin(\theta)}{5}+\frac{W \times cos(\theta)}{2}$\\ 
\hline
\end{tabular}}
\end{center}
    
\caption{ROI's Cartesian coordinates}
\label{tab1:coordinates}
\end{table}

The 3-D tensor state representation contains three 2-D grids where the first 2-D grid represents the cells that are occupied by the ego vehicle and surrounding pedestrians. Relative speed and relative heading direction of the corresponding pedestrians with respect to the ego vehicle are captured in the second and the third 2-D grids, respectively. Speed is expressed in \textit{meters per second (m/s)}, and heading direction in \textit{degrees}. 

We assume that the length and the width of the ego vehicle are $4.5m$ and $2.0m$, respectively. The ROI's parameters are selected as $L=20m$, $W=15m$, $l=0.25m$ and $w=0.25m$. As a result, each 2-D grid is an $80 \times 60$ matrix. The ego vehicle's center of gravity occupies cell $(64,15)$ of the first 2-D grid. 

\subsubsection{Action Space}
\label{sec4.B.2}

We control the ego vehicle by adjusting the throttle value $a \in [-1,1]$. The action space includes four discrete actions presented in Table~\ref{tab2:action space}.

\begin{table}
\begin{center}
\scalebox{0.9}{
\begin{tabular}{ |c|c|c|c| } 
\hline
Action & Throttle Value & Description \\
\hline
$a_{0}$  & $-1.0$ & full brake\\
$a_{1}$  & $-0.4$ & decelerate\\
$a_{2}$  & $+0.2$ & $accelerate_{1}$\\
$a_{3}$  & $+1.0$ & $accelerate_{2}$\\
\hline
\end{tabular}}
\end{center}
\caption{Action Space}
    \label{tab2:action space}
\end{table}

\subsubsection{Reward Function}
\label{sec4.B.3}
The goal of the ego vehicle is to make a left-turn within a limited time frame ($t=45$ sec) while avoiding collisions with any pedestrians and maintaining a speed below the speed limit ($10$ m/s). To this end, we designed a conditional reward function inspired by~\cite{everett2019collision} where the switching criteria are based on the distance of the ego vehicle's right and left bumpers to the closest pedestrians denoted by $d_{right}$ and $d_{left}$, respectively. Therefore, at each time step, if both distances are greater than a threshold $d_{1}$, the ego vehicle receives rewards only based on its distance to the target position ($X_{t}$) and its velocity ($V$). Otherwise, the ego vehicle receives rewards according to $d_{right}$ and $d_{left}$ as depicted in Figure~\ref{fig2:rewardfunction}. The hyperparameters of the reward function are shown in Table~\ref{tab3:rewardfunction}. Since the reward function is designed such that the agent receives the highest reward at the end of the navigation task, it is not necessary to assign excessive reward when the agent reaches the goal. The ego vehicle reaches a terminal state when it completes the left-turn, collides with a pedestrian, or it runs out of time, and then the episode restarts. The total amount of reward that the ego vehicle receives for every episode is the summation of all $reward$ over all of the time steps. 

\begin{table}
\begin{center}
\scalebox{0.9}{
\begin{tabular}{ |c|c|c|c|c| } 
\hline
Hyperprameter & Value & Hyperprameter & Value \\
\hline
$d_1$ & $7(m)$ &$r_2$ & $+0.005$\\ 
$d_2$ & $25(m)$&$r_3$ & $-0.25$ \\ 
$d_3$ & $1(m)$ &$r_4$ & $+0.25$\\ 
$d_4$ & $2(m)$ &$r_5$ & $-5.0$\\ 
$v_1$ & $1.5 (m/s)$& $r_6$ & $-1.5$\\
$v_2$ & $10 (m/s)$& $r_7$ & $-0.25$\\
$r_1$ & $+0.005$ & &\\

\hline
\end{tabular}}
\end{center}
\caption{Reward function hyperparameters}
    \label{tab3:rewardfunction}

\end{table}

\section{Experiments}
\label{sec5}
\begin{figure}[t]
    \centering
    \includegraphics[scale = 0.5]{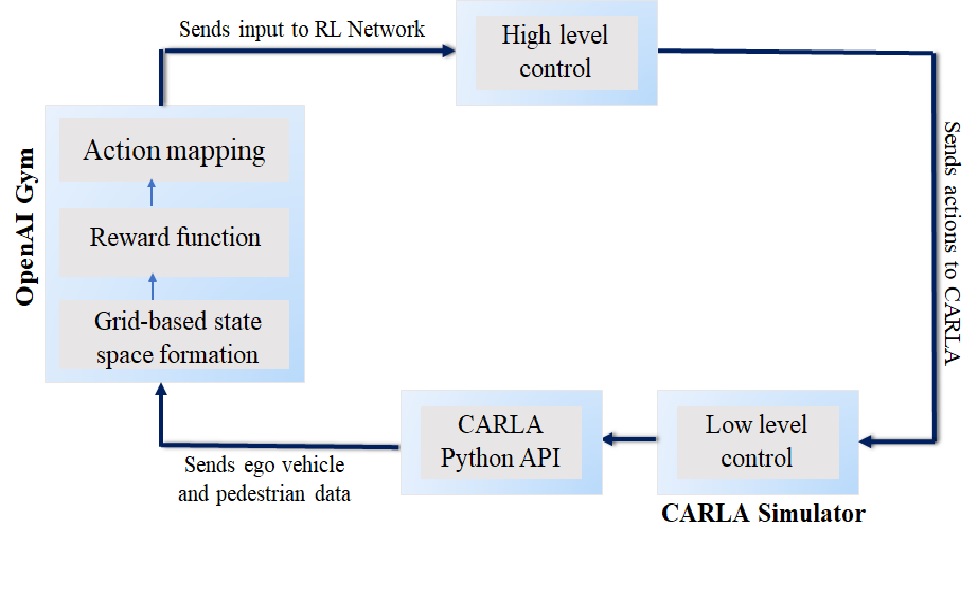}
    \caption{Decision-making process architecture.}
    \label{fig3:Decision-making}
\end{figure}
\subsection{Simulation Setup}
\label{sec5.A}
The system consists of the four components shown in Figure~\ref{fig3:Decision-making}. High-level control (a throttle value) is generated by the method and low-level control is executed by CARLA~\cite{dosovitskiy2017carla}, which is an open-source simulator for autonomous driving scenario development. The ego vehicle and pedestrian data generated by the CARLA API are then fed into the interface component of the OpenAI Gym toolkit~\cite{brockman2016openai}. OpenAI Gym is an open-source library for developing reinforcement learning algorithms. At every time step, this module calculates the grid-based state space representation, the assigned rewards to the ego vehicle, and action mapping as discussed in Section~\ref{sec4.B}. To complete the control loop, a fourth component, the reinforcement learning module constructed using the Tensorflow library selects the best possible discrete action and returns it to the simulation.

\subsection{Reinforcement Learning Architecture}
\label{sec5.B}

We examine and compare the performance of three different reinforcement learning methods for our problem. 
\subsubsection{DQN Method}
\label{sec5.B.1}
\begin{figure}[t]
    \centering
    \includegraphics[scale = 0.55]{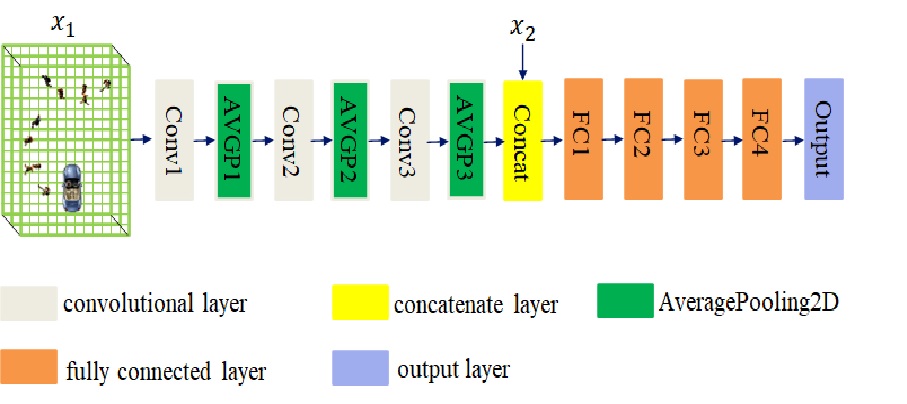}
    \caption{Deep Q-network architecture.}
    \label{fig4:DQN}
\end{figure}

A DQN architecture is shown in Figure~\ref{fig4:DQN}. The input $x_1$ to the DQN is the 3-D tenor discussed in Section~\ref{sec4.B.1} and the input $x_2$ is the ego vehicle velocity. Three convolutional layers followed by an ReLU activation function and an Averagepooling2D layer are used to extract the low-level features of the 3D-tensor. The output of the convolutional layers is then fed into a flatten layer and is concatenated with the ego vehicle velocity. The new vector is then propagated through four fully connected layers. The parameters of the network are described in Table~\ref{tab3:DQN}.

\begin{table}
\begin{center}
\scalebox{0.7}{
\begin{tabular}{ |c|c|c|c|c|c|c|c| } 
\hline
Layer & Activation Function & Filters & Units & Kernel Size & stride  \\
\hline
Conv1 & ReLU & $64$ & - & $3\times 3$ & $1 \times 1$  \\
AVGP1 & - & $1$ & - & $5\times 5$ & $3 \times 3$  \\
Conv2 & ReLU & $64$ & - & $3\times 3$ & $1 \times 1$  \\
AVGP2 & - & $1$ & - & $5\times 5$ & $3 \times 3$  \\
Conv3 & ReLU & $64$ & - & $3\times 3$ & $1 \times 1$  \\
AVGP3 & - & $1$ & - & $5\times 5$ & $3 \times 3$  \\
FC1 & ReLU & - & 512 & - & -  \\
FC2 & ReLU & - & 256 & - & -  \\
FC3 & ReLU & - & 64 & - & -  \\
FC4 & ReLU & - & 4 & - & -  \\
\hline
\end{tabular}}
\end{center}
\caption{Parameters of the RL architecture.}
\label{tab3:DQN}
\end{table}

\subsubsection{DDQN Method}
\label{sec5.B.2}
For the DDQN, the action selection and action evaluation tasks are handled with two separate networks, the online network and the target network. Therefore, a DDQN consists of two networks where each one shares the DQN structure. The target network parameters are updated after 5000 time steps using the latest parameters of the online network.   

\subsubsection{DDQN/PER Method}
\label{sec5.B.3}

This approach combines the DDQN method and PER implementation. To conduct a fair comparison, we use exactly the same DDQN architecture as the previous Section~\ref{sec5.B.2}.  

\subsection{Networks Training}
\label{sec5.C}

To generate training data, the unsignalized intersection left turn scenario was constructed and simulated in CARLA. The networks were trained for 450 episodes. Each step generated a transition and the experience replay memory buffer stored up to 10,000 transitions. The learning process began when the experience replay memory reached a threshold of 750 transitions. The DQN and DDQN methods randomly selected a mini-batch of size 32 transitions from the replay memory and then updated the weights of the networks accordingly. However, the DDQN/PER approach instead used a stochastic sampling method (Eq. 9) to choose the same size mini-batch of data for training its network. All methods used the RMSprop optimization~\cite{karparthy2017peek} algorithm with a learning rate of 0.00025. 

During the training process, a discount factor of 0.95 was applied to discount future rewards. The $\epsilon$ greedy approach was followed in order to allow the networks to explore at each time step, an action was chosen randomly with a probability of $\epsilon$, otherwise, an action is selected by the networks. The value $\epsilon$ was initialized to 1. It annealed to a minimum value of $0.05$ using the decay value of $0.99$ in order to limit the exploratory behavior towards the end of the training process. For the DDQN/PER method, the prioritization rate $\alpha$ was set to $0.6$ and the compensation rate $\beta$ was initialized to $0.4$ linearly annealing to 1.     

\subsection{Scenario}
\label{sec52.C}

The simulator ran on synchronous mode (15 FPS) which involved a constant time interval between each step taken by the ego vehicle in the simulation. At the beginning of every episode, the ego vehicle and a random number of pedestrians (between five and thirty) were spawned at an unsignalized intersection as shown in Figure~\ref{fig1:intersection}. Five more pedestrians were spawned every ten seconds after the start of the episode. The pedestrians moved to random destination points with a random velocity between $0.2-1.8$ m/s. A trajectory was defined for the ego vehicle by automatically placing a hundred waypoints between the start point to the endpoint. The goal of the ego vehicle was to make a left turn within the specific time while avoiding collision with any pedestrian and not violating the speed limit. Forty-five seconds were allotted for each episode. The speed limit for the ego vehicle was $10$ m/s. At each time step, the action generated by the network was executed in the simulator and the network was trained. The ego vehicle reached the terminal state when it either completes the left-turn, collides with a pedestrian, or runs out of time, at which point the episode restarts.

\section{Results and Discussion}
\label{sec6}
\begin{figure*}[t]
    \centering
    \includegraphics[scale = 0.55]{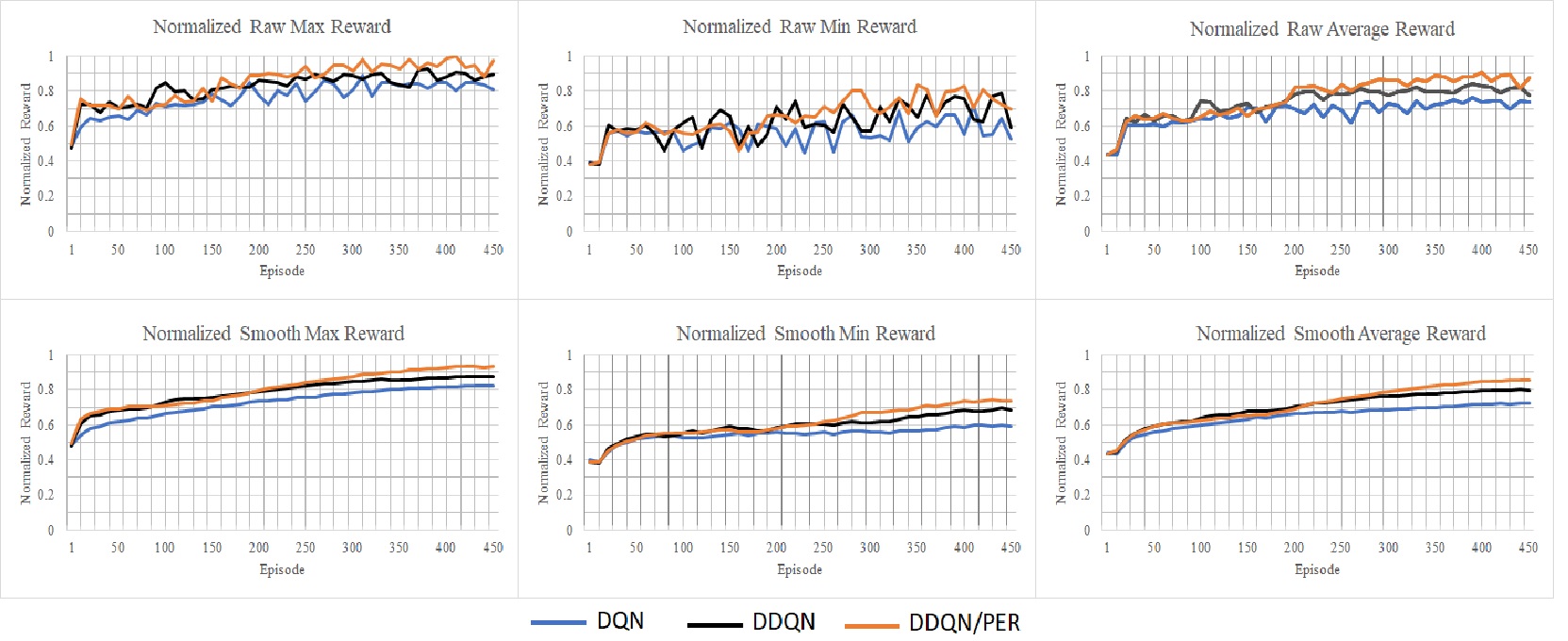}
    \caption{RL networks training performance comparison.}
    \label{fig5:Results_train}
\end{figure*}

\begin{table}
\caption{Comparison of test performance results}
\label{tab5:testresults}
\bigskip
\scalebox{1.0}{
\begin{tabular}{@{}*{3}{c}@{}}
\multicolumn{3}{@{}l}{\em(a) The intersection that the DRL networks are trained on}\\
\toprule
method & Collision-free &Successful \\
& episodes& episodes \\
\midrule
Rule-based & 38\%& 38\% \\                                                     
DQN & 99\%& 82\% \\                                                               
DDQN & \textbf{100}\%& 92\%\\                                                         
DDQN/PER & \textbf{100}\%& \textbf{93}\% \\                          

\bottomrule
\end{tabular}}

\bigskip
\scalebox{1.0}{
\begin{tabular}{@{}*{3}{c}@{}}
\multicolumn{3}{@{}l}{\em(b) An unseen intersection with a different topology}\\
\toprule
method & Collision-free &Successful\\
& episodes& episodes \\
\midrule
Rule-based & 35\%& 35\% \\                                                    
DQN & 99\%&  85\% \\                                                               
DDQN & \textbf{100}\%&  92\%\\                                                         
DDQN/PER & \textbf{100}\%&  \textbf{94}\%\\                          

\bottomrule
\end{tabular}}
\end{table}

The training performance of the three type of networks on the unsignalized intersection task is compared in Figure~\ref{fig5:Results_train}. The maximum reward, the minimum reward, and the average reward of 10 episodes are compared and the depicted rewards are normalized and smoothed. Due to substantial overestimation associated with the DQN~\cite{van2015deep}, it was outperformed by the other two networks. Because the DDQN/PER employs a stochastic sampling method instead of random sampling to select the training batch, it trains the most efficiently and exhibits the highest normalized maximum, minimum and average reward during the training phase. At the end of the training process, the average normalized reward for the DDQN/PER was $0.07$ and $0.13$ greater than that of the DDQN and DQN, respectively. The ascending trend in the normalized maximum, minimum, and average rewards for all networks suggests that the networks are improving. To evaluate the system's learning, the ability of each network to navigate the intersection without speeding and timing out was examined. 

\begin{figure}[t]
    \centering
    \includegraphics[scale = 0.5]{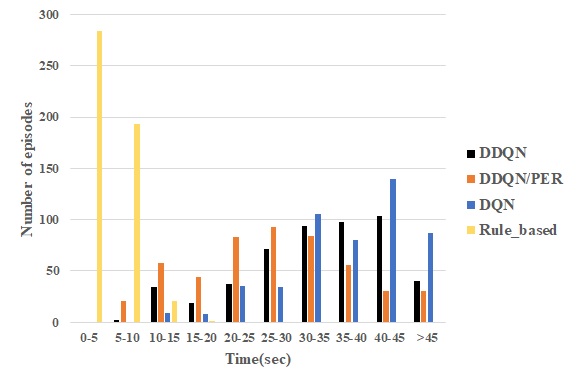}
    \caption{Histogram plot of average intersection crossing time for the rule-based and RL methods.}
    \label{fig5:Results_train_time}
\end{figure}

\begin{figure}[t]
    \centering
    \includegraphics[scale = 0.5]{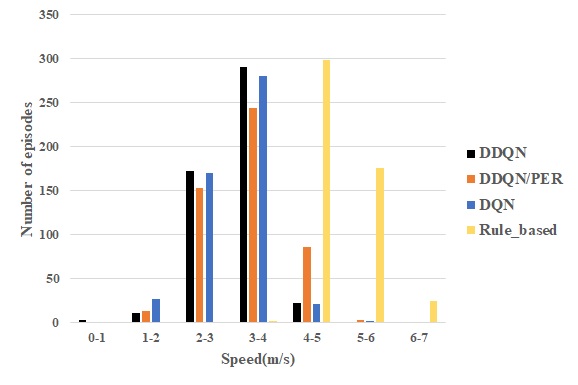}
    \caption{Histogram plot of average intersection crossing speed for the rule-based and RL methods.}
    \label{fig5:Results_train_speed}
\end{figure}

\begin{figure}[t]
    \centering
    \includegraphics[scale = 0.5]{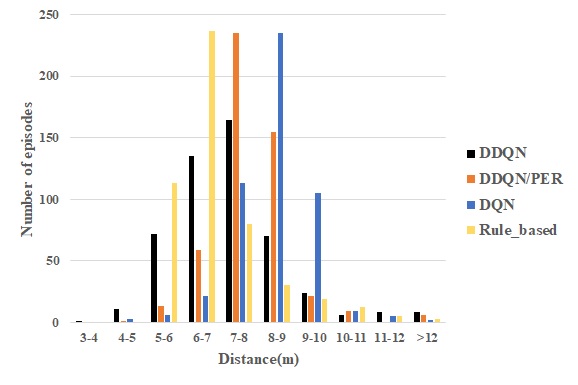}
    \caption{Histogram plot of average distance from the closest pedestrian for the rule-based and RL methods.}
    \label{fig5:Results_train_distance}
\end{figure}
The different method's performance was compared in two experiments. Each experiment included 250 episodes and used the same definition of the terminal state as used for the training process. The same method for spawning the pedestrians at the target unsignalized intersection was followed. Both experiments compared our methods to a rule-based method. The rule-based method is provided by CARLA which executes a policy based on a hand-engineered strategy for navigating around pedestrians. The policy mainly relies on a time to collision approach to decide when to cross the intersections. 

In the first experiment, these methods were tested on exactly the same unsignalized intersection on which the networks were trained. For the second experiment, the networks were tested on an unseen unsignalized intersection with different topology. The networks were trained on a $25 \times 25$ three-way unsignalized intersection while the unseen intersection was a $26 \times 17$ four-way unsignalized intersection. For each experiment, the following metrics were used to evaluate the methods' performance: the percentage of collision-free and successful episodes. The successful episodes are the episodes that the autonomous vehicle neither had a collision with pedestrians, nor ran out of time, nor violated the speed limit. Moreover, we compared the histogram plots of average intersection crossing time (sec) (Figure~\ref{fig5:Results_train_time}), average interaction crossing speed (m/s) (Figure~\ref{fig5:Results_train_speed}), and average distance of ego vehicle from the closest pedestrian (m) (Figure~\ref{fig5:Results_train_distance}) for the RL methods.   

The results for the two experiments are presented in Table~\ref{tab5:testresults}. During both experiments (according to the histogram plots), although the average speed and the average intersection crossing time for the rule-based method outperforms our methods, 62\% and 65\% of the episodes result in collisions with pedestrians for experiment one and experiment two, respectively. Given a success rate of only $35\%-40\%$ success rate for completing the left turn, the rule-based method is not a viable solution to this problem. Compared to a rule-based approach, our methods clearly show better performance with $99\%$, $100\%$ and $100\%$ collision-free episodes for the DQN, DDQN, and DDQN/PER methods, respectively. Qualitatively, we see that these methods learn to decrease their speed and wait for pedestrians to cross in order to avoid a collision with a pedestrian. These methods are also more successful at completing the navigation task than rule-based methods. We therefore conclude that our methods are safer then the standard rule-based method. 

According to the histogram plots and their average value for intersection crossing time (sec), interaction crossing speed (m/s), and distance of ego vehicle from the closest pedestrian (m), among the three methods, similar to the training results, DDQN/PER outperforms the other two. This method has the fastest average intersection crossing time ($30.08$ sec for experiment one and $27.86$ sec for experiment two), the greatest average speed ($3.38$ m/s for experiment one and $3.35$ m/s for experiment two), and the greatest distance from the closest pedestrian ($7.8$ m for experiment one and $6.9$ m for experiment two) as depicted in Table~\ref{tab5:testresults}. Overall, the DDQN and DDQN/PER methods are capable of safely navigating through crowds and following the traffic rules for different unsignalized intersections after being trained on one topology and then tested on another topology. 

\begin{table*}
\begin{center}
\scalebox{0.85}{
\begin{tabular}{ |l|l|l|l|l| } 
\hline
Authors & Results & Limitations & Scenario & Simulation Environment \\
\hline
Deshpande et al.~\cite{deshpande2019deep} &  No collision &\makecell[l]{Did not Use a high-fidelity autonomous \\ driving simulation and the method was \\ not tested in a pedestrian-rich environment.} &
\makecell[l]{Intersection crossing among \\only two pedestrians.} & SUMO~\cite{lopez2018microscopic} \\ 
\hline

Bouton et al.~\cite{bouton2019safe} & No collision & \makecell[l]{Used a very simplistic simulated environment\\ and the computation time for this method grows \\drastically with the number of pedestrians around \\ the ego vehicle.} & \makecell[l]{Intersection crossing (left turn) \\ among pedestrians and cars.} &
\makecell[l]{They created their \\own simulated environment.}  \\ 
\hline
Deshpande et al.~\cite{deshpande2020behavioral} &  70\% collision-free & \makecell[l]{Did not generate a safe behavior for navigating\\ around pedestrians.} &\makecell[l]{Intersection crossing when\\ the pedestrians might jaywalk.} & CARLA \\
\hline
\end{tabular}}
\end{center}
\caption{The related work and the summary of approach for the problem of autonomous vehicle navigation among pedestrians. Unfortunately, because different authors have used different simulation environments and different actions spaces, direct comparison of our approach to these prior methods is not feasible.}
    \label{tab3:relatedwork}

\end{table*}

The related work and their limitations for the problem of autonomous vehicle navigation among pedestrians are presented in Table~\ref{tab3:relatedwork}. Prior approaches to this problem that utilized deep reinforcement learning either have not resulted in a collision-free control of the vehicle, or have not been tested in a pedestrian-rich environment in a high-fidelity autonomous driving simulation, or they are not suitable for the real-time applications. Our method shows that by employing the 3D state space representation of the environment and our innovative conditional reward function combined with the DDQN and DDQN/PER training methods, to the best of our knowledge, we have developed the reinforcement learning-based decision-making process for an AV that is not only safe ($100\%$ collision-free episodes) but also capable of successfully ($92-94\%$ completed navigation tasks episodes) navigating at unisignalized intersections in a pedestrian-rich urban environment regardless of the intersection topology. Moreover, due to the small computation time required by the networks, we believe that these methods can be used in real-time.

\section{Conclusion}
\label{sec7}

This paper has examined three different deep reinforcement learning approaches for controlling an autonomous vehicle as it navigates through an unsignalized intersection crowded with pedestrians. Of the three approaches, the DDQN/PER outperforms the other two methods. Comparing these deep reinforcement learning methods to a rule-based method, given the 3D state space representation and our innovative conditional reward function, we find that the methods: 1) drastically increase safety during navigation, 2) are much better at obeying the speed limit and 3) achieve much higher success rate in a navigation task. We also consider how these methods trained on one intersection topology perform on another, different topology. The results reveal that our methods maintain almost the same performance. 

We have assumed that the environment is fully observable and the pedestrian information collected by the AVs sensors is accurate. Our future work will explore the use of safe deep reinforcement learning (SDRL) approaches for the same scenario using a partially observable Markov decision process (POMDP) formulation.






{\small
\bibliographystyle{IEEEtran.bst}
\bibliography{bibliography.bib}
}

\end{document}